\documentclass{article}
\usepackage{spconf,amsmath,graphicx}
\usepackage{subfig}
\usepackage[T1]{fontenc}
\usepackage{bbding}
\usepackage[colorlinks,linkcolor=red,anchorcolor=green,citecolor=blue]{hyperref}
\usepackage{color,xcolor}
\newcommand\Mark[1]{\textsuperscript#1}
\usepackage[noend]{algpseudocode}
\usepackage{algorithmicx,algorithm}

\usepackage{arydshln} 

\newcommand{\Ree}[1]{\textcolor{black}{#1}}
\newcommand{\Rone}[1]{\textcolor{black}{#1}}
\newcommand{\R}[1]{\textcolor{black}{#1}}
\usepackage[inline]{enumitem}
\usepackage{graphicx}
\usepackage{epstopdf}

\newif\ifarxiv
\arxivtrue
\usepackage{lastpage}
\usepackage{fancyhdr}
\fancypagestyle{firstpage}
{
  \fancyhf{}
  
  \ifarxiv
  \fancyfoot[C]{\scriptsize \copyright~2023 IEEE. Personal use of this material is permitted. Permission from IEEE must be obtained for all other uses, in any current or future media, including reprinting/republishing this material for advertising or promotional purposes, creating new collective works, for resale or redistribution to servers or lists, or reuse of any copyrighted component of this work in other works. 
  This work has been accepted at 2023 IEEE International Conference on Acoustics, Speech, and Signal Processing (ICASSP 2023).}
  \else
    
  \fi
}

\title{BEBERT: Efficient and Robust Binary Ensemble BERT}
\name{}
\address{}
\name{Jiayi Tian\Mark{1}, Chao Fang\Mark{1}, Haonan Wang\Mark{2}, and Zhongfeng Wang\Mark{1}\sthanks{Corresponding author. This work was supported in part by the National Natural Science Foundation of China under Grant 62174084, 62104097, in part by the High-Level Personnel Project of Jiangsu Province under Grant JSSCBS20210034, in part by the Key Research Plan of Jiangsu Province of China under Grant BE2019003-4, and in part by Postgraduate Research \& Practice Innovation Program of Jiangsu Province under Grant No. 149.}}
\address{\Mark{1}Nanjing University, Nanjing, China~~~~~~\Mark{2}University of Southern California, Los Angeles, USA}

\begin{document}
%
\maketitle
\thispagestyle{firstpage}
%
%

\begin{abstract}

Pre-trained BERT models have achieved impressive accuracy on natural language processing (NLP) tasks. \R{However, their} excessive amount of parameters hinders them from efficient deployment on edge devices.
Binarization of the BERT models can significantly alleviate \R{this issue but comes} with a severe accuracy drop compared with their full-precision counterparts.
In this paper, we propose an efficient and robust binary ensemble BERT (BEBERT) to bridge the accuracy gap.
To the best of our knowledge, this is the first work employing ensemble techniques on binary BERTs, yielding BEBERT, which achieves superior accuracy while retaining computational efficiency.
Furthermore, \R{we remove the knowledge distillation procedures during ensemble to speed up the training process without compromising accuracy.}
Experimental results on the GLUE benchmark show that the proposed BEBERT significantly outperforms the existing binary BERT models in accuracy and robustness with a 2× speedup \R{on training time}. 
Moreover, our BEBERT has only a negligible accuracy loss of 0.3\% compared to the full-precision baseline while saving 15× and 13× in FLOPs and model size, respectively.
In addition, BEBERT also outperforms other compressed BERTs in accuracy by up to 6.7\%.
\end{abstract}
\vspace{-0.5em}
\section{Introduction}

Pre-trained BERT models \cite{devlin2019bert} \Rone{achieve outstanding accuracy on various} natural language processing (NLP) tasks. 
However, deploying them on resource-\Rone{constrained edge} devices is \Rone{difficult} due to \Rone{their} massive parameters and floating-point operations (FLOPs). 
To tackle the challenges above, model compression techniques, including quantization \cite{zafrir2019q8bert, shen2020q, boo2020fixed, kim2021bert}, pruning \cite{chen2021earlybert, xie2021elbert, hou2020dynabert, fang2022algorithm}, and knowledge distillation (KD) \cite{jiao2020tinybert, sun2020mobilebert, wang2020minilm, sanh2019distilbert}, are widely studied and applied for deploying BERTs in resource-constrained and real-time scenarios.

\Rone{Among them, quantization, which utilizes lower bit-width representation for model parameters, emerges as an efficient way to deploy compact models on edge devices.}
Binarization \cite{sakr2018true, zhang2022dynamic, xing2022binary}, the most aggressive quantization scheme, can significantly reduce storage consumption and inference cost by quantizing parameters to 1-bit.
\Rone{However, existing binary BERTs \cite{bai2021binarybert, qin2022bibert} suffer from a severe accuracy drop and unstable issues, making them impractical for deployment.}
\Rone{For instance, for the CoLA task on the GLUE benchmark \cite{wang2018glue}, the accuracy of BinaryBERT drops by 11.3\%, and BiBERT fails to attain half of the accuracy compared to the full-precision BERT.}
\Rone{Furthermore, \cite{bai2021binarybert} points out binary BERT models are unstable to the input perturbations and hence suffer from severe robustness issues, leading to limited generalization capabilities.}
Additionally, binary BERTs require a long training time since time-consuming KD \cite{hinton2015distilling} methods are required to improve their performance.
\Rone{In summary, existing binary BERT models struggle with (1) impractical model accuracy, (2) fragile model robustness, and (3) expensive training costs.}

To address the above issues, \R{we employ ensemble learning on binary BERT models, yielding Binary Ensemble BERT (BEBERT)\footnote{Available at \href{https://github.com/TTTTTTris/BEBERT}{https://github.com/TTTTTTris/BEBERT}.}, which achieves superior accuracy while retaining computational efficiency.}
Moreover, \R{we accelerate the training process of BEBERT without impact on model accuracy by removing the time-consuming KD procedures.}
Experiments have been conducted on multiple NLP tasks on the GLUE benchmark to evaluate the accuracy, robustness, and training efficiency of BEBERT. The results demonstrate the great potential of BEBERT for edge deployment.
The main contributions of this paper can \Rone{be summarized} as follows: \\
\begin{figure}[htbp]  
\centering  
\includegraphics[width=0.9\linewidth,scale=1.00]{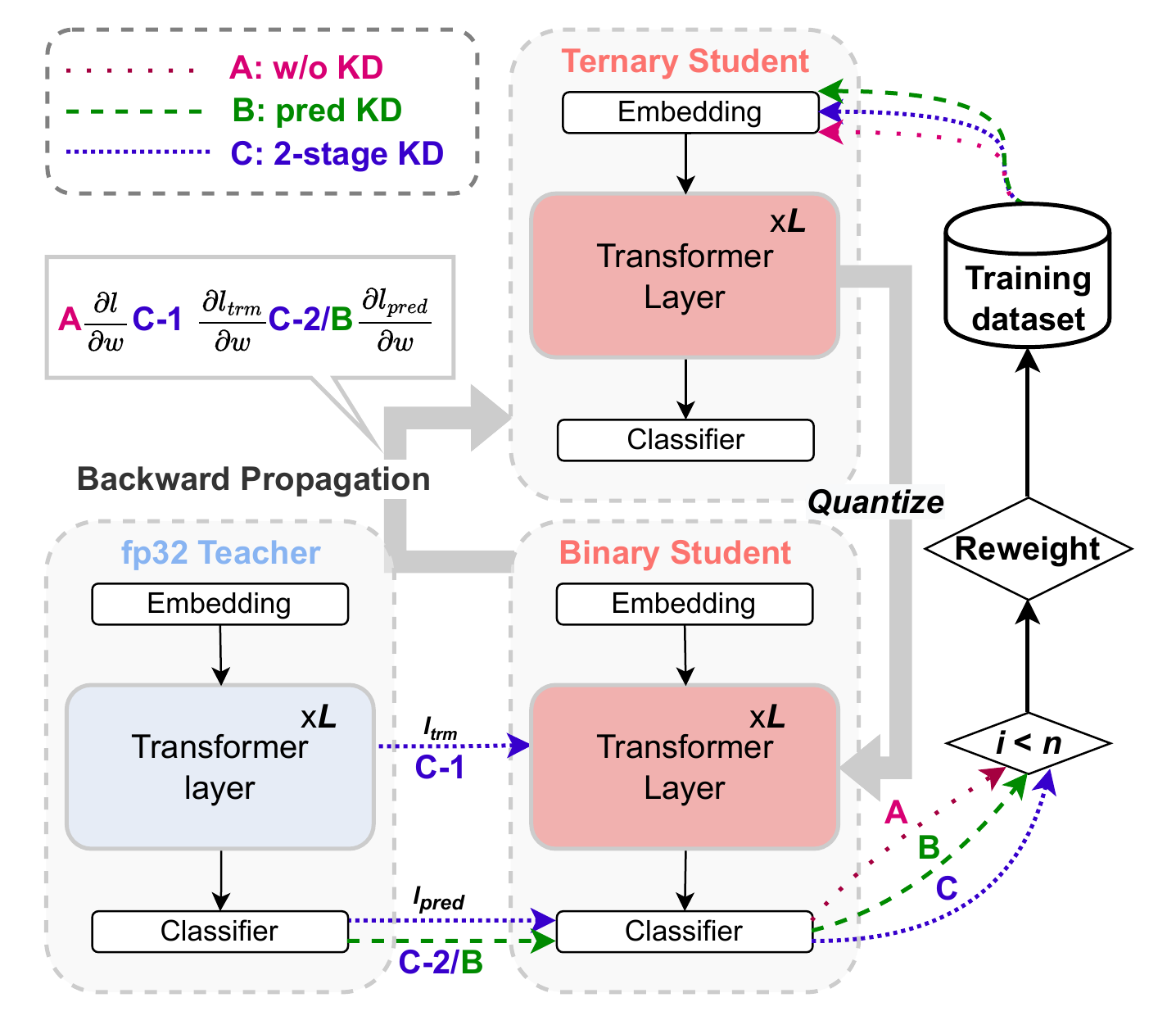}
\caption{Structure of BinaryBERT-based BEBERT. The dashed lines denoted with A, B, and C represent combining \R{ensemble} with different KD \R{strategies}.}
\label{fig:1}
\vspace{-1em}
\end{figure}\indent 
\begin{enumerate*}[label =\roman*)]
    \item To the best of our knowledge, we \R{are the first to} introduce ensemble learning to binary BERT models to improve accuracy and robustness. 
    \Rone{To illustrate the generalization capability, \R{the} ensemble method is conducted on both BinaryBERT and BiBERT, improving accuracy by \textbf{2.4\%} and \textbf{4.0\%}, respectively. \R{It also reduces} the variance of output accuracy by \textbf{65\%} and \textbf{61\%}, respectively.}\\
    \indent \item \R{We remove the KD procedures during ensemble to accelerate the training process.} Experiments show that our BinaryBERT-based BEBERT w/o KD achieves nearly \textbf{2×} speedup \R{on training time} compared to the BinaryBERT.\\
    \indent \item 
    \Rone{The proposed BEBERT enables practical deployment of binary BERT models on resource-limited devices.}
    \Rone{Experimental results show that BinaryBERT-based BEBERT \R{has only a negligible accuracy drop of \textbf{0.3\%} compared to the full-precision baseline while saving \textbf{15×} and \textbf{13×} in FLOPs and model size, respectively.} \R{Moreover, it} outperforms the state-of-the-art (SOTA) compressed BERTs \cite{hou2020dynabert, jiao2020tinybert, sanh2019distilbert, wang2020minilm, chen2021earlybert} in accuracy by up to \textbf{6.7\%}.}
\end{enumerate*}

\vspace{-0.6em}
\section{Related Works} \label{sec:rel}

\textbf{Quantization} \cite{zafrir2019q8bert, shen2020q, kim2021bert, bai2021binarybert, zhang2020ternarybert, qin2022bibert}, as an efficient way to obtain a compact model, has been widely exploited for BERT deployment.
Q8BERT \cite{zafrir2019q8bert} and I-BERT \cite{kim2021bert} achieve 8-bit integer quantization for parameters of BERTs at first.
Moreover, Q-BERT \cite{shen2020q} implements a low-bit representation of BERT with a hybrid quantization scheme, and TernaryBERT \cite{zhang2020ternarybert} achieves 2-bit quantization using ternary parameters.
\Ree{Furthermore, BinaryBERT \cite{bai2021binarybert} achieves memory efficiency with 1-bit weights with 4-bit activations while achieve the highest accuracy among all binary BERTs. And BiBERT \cite{qin2022bibert} is the first fully binarized model with improved accuracy.}
However, existing binary BERTs still suffer from a severe accuracy drop, making it impractical for model deployment.
To bridge this gap, our BEBERT integrates several binary BERTs to attain a comparable accuracy to the full-precision BERT, with significant savings on both FLOPs and model size.

\Rone{\textbf{Other compression methods} \cite{hou2020dynabert, jiao2020tinybert, wang2020minilm, sanh2019distilbert, chen2021earlybert} beyond quantization have been exploited for efficient BERTs as well.}
\Rone{DistilBERT \cite{sanh2019distilbert} and miniLM \cite{wang2020minilm} utilize knowledge distillation (KD) \cite{hinton2015distilling} to decrease the depth of BERT, which significantly reduces computational complexity.}
\Rone{Based on the KD, TinyBERT \cite{jiao2020tinybert} introduces a novel data augmentation (DA) method to improve model accuracy.} DynaBERT \cite{hou2020dynabert} effectively reduces the number of operations by pruning and parameter sharing and EarlyBERT \cite{chen2021earlybert} further improves the training efficiency based on structured pruning.
\Rone{In contrast, our BEBERT integrates multiple binary BERTs using ensemble learning, which accelerates the training process and outperforms these compressed BERTs in terms of model size and accuracy.}

\textbf{Ensemble learning} \cite{zhou2021ensemble} integrates the predictions of several weak classifiers to improve model accuracy.
\Ree{Although neural networks (NNs) are neither weak nor unstable classifiers, prior works \cite{mocerino2020fast, gao2022soft, ponzina20212, zhu2019binary} successfully used ensemble methods to improve the performance of NNs. This indicates the potential for using ensemble techniques to enhance the performance of binary BERT models. Two common ensemble techniques are boosting \cite{freund1996experiments} and bagging \cite{breiman1996bagging}.
Theoretically, bagging is thought only to reduce the variance while boosting can reduce both bias and variance. Specifically, bias portrays the fitting ability of the learning algorithm itself, and variance portrays the effect caused by the data perturbation.
Therefore, focusing on increasing the accuracy, we choose AdaBoost \cite{freund1996experiments} as the main ensemble method.}
Our BEBERT makes the first attempt to leverage ensemble techniques on binary BERTs to improve model accuracy and retain computational efficiency, which can be easily integrated with existing pipelines for efficient binary BERT.
\vspace{-0.5em}
\section{Methods} \label{sec:methods}

\subsection{BEBERT Overview}
Fig. \ref{fig:1} shows the architecture of BEBERT based on BinaryBERT \cite{bai2021binarybert}, which is integrated by BinaryBERT models using AdaBoost \cite{freund1996experiments}. 
\Ree{We first assign sample weights to each training sample and then update them in each iteration focusing on the wrongly predicted elements.}
\Ree{During the forward propagation, we train a ternary student and quantize the weights to 1 bit and the activations to 4 bits. When KD is used, the loss is computed between the binary student and the full-precision teacher, or else with the labels. We subsequently update the weights of the ternary student using the backward gradients.}
The training process of BEBERT based on BiBERT is similar to that based on BinaryBERT, except that BiBERT is quantized from a full-precision student and distilled by the DMD method \cite{qin2022bibert}.
Note that the original two-stage KD \cite{jiao2020tinybert} contains distillation for Transformer layers and the prediction layer, introducing extra forward and backward propagation (BP) steps in training.
Therefore, we propose to either distill the prediction layer or remove the KD procedures to reduce the training costs.

\begin{algorithm}[htbp]
\small
\caption{\Rone{Training Process of BEBERT}} 
\label{agr: 1}
\begin{algorithmic}[1]
\State \textbf{Initialize:} $m$ examples $\langle(x_1, y_1),...,(x_m,y_m)\rangle$ in the training set; $D_1=(w_{11}...w_{1m})=(\frac{1}{m}...\frac{1}{m})$ is the sample weight;
\State \textbf{Input:} training set with sample weights $D_1$;
\For{$i = 1$ to $N$}
    \For{$iter = 1$ to $T$} 
        \State Get next mini-batch of training data;
        \State Binarize $w$ in student model to $w_b$;
        \If {two-stage KD} 
            \State Compute $L_{trm}$ between $h^T$ and $h_i^S$; BP;
            \State Compute $L_{pred}$ between $h^T$ and $h_i^S$; BP;
        \ElsIf {prediction KD}   
            \State Compute $L_{pred}$ between $h^T$ and $h_i^S$; BP;
        \Else
            \State Compute loss between $Y$ and $h_i$; BP;
        \EndIf
        \State Update weights with the gradients $\frac{\partial L}{\partial w_b}$;
    \EndFor
    \State Calculate the error $e_i$ and weight $\alpha_i$ of $h_i^S$; 
    \State Update the sample weight $D_{i+1}$; 
\EndFor
\State \textbf{Output:} The final hypothesis $H\leftarrow \Sigma_{i=1}^N\alpha_i h_i(x_i)$; 
\end{algorithmic}
\end{algorithm}

\vspace{-0.5em}

\subsection{Training Process of BEBERT}
\label{sec:ensemble-learning}
\Ree{In the training process, we use AdaBoost \cite{freund1996experiments} to integrate multiple binary BERTs focusing on the accuracy increase.}
The detailed training process is shown in Algorithm~\ref{agr: 1}.
The AdaBoost-based BEBERT takes a training set $S$ of $m$ examples $\langle(x_1, y_1),...,(x_m,y_m)\rangle$ as input, where $y_j\in Y$ represents the label of $j$-th sample.
\Ree{During training}, the boosting algorithm calls the \Ree{ternary model} to train for $N$ rounds, generating a binary \Ree{classifier $h_i$ (or $h_i^S$ when KD is employed)} in each round.
In the $i$-th round, AdaBoost provides the training set with a distribution $D_i$ as the sample weight; The initial distribution $D_1$ is uniform over $S$.
\Ree{We calculate an error $e_i$ and a weight $\alpha_i$ for each classifier $h_i$, and then updates the sample weight $D_{i+1}$ in each iteration} focusing on minimizing the error $e_i=P_{j\sim D_i}(h_i(x_j)\neq y_j)$.
At last, the booster combines the weak hypotheses into a single final hypothesis $H\leftarrow \Sigma_{i=1}^N\alpha_i h_i(x_i)$.

\vspace{-0.5em}
\subsection{Removing KD for Efficient Training}
\label{sec: KD}
\begin{figure}[htbp]  
\centering  
\subfloat[CoLA]{  
\centering  
\includegraphics[width=0.5\linewidth,scale=1.00]{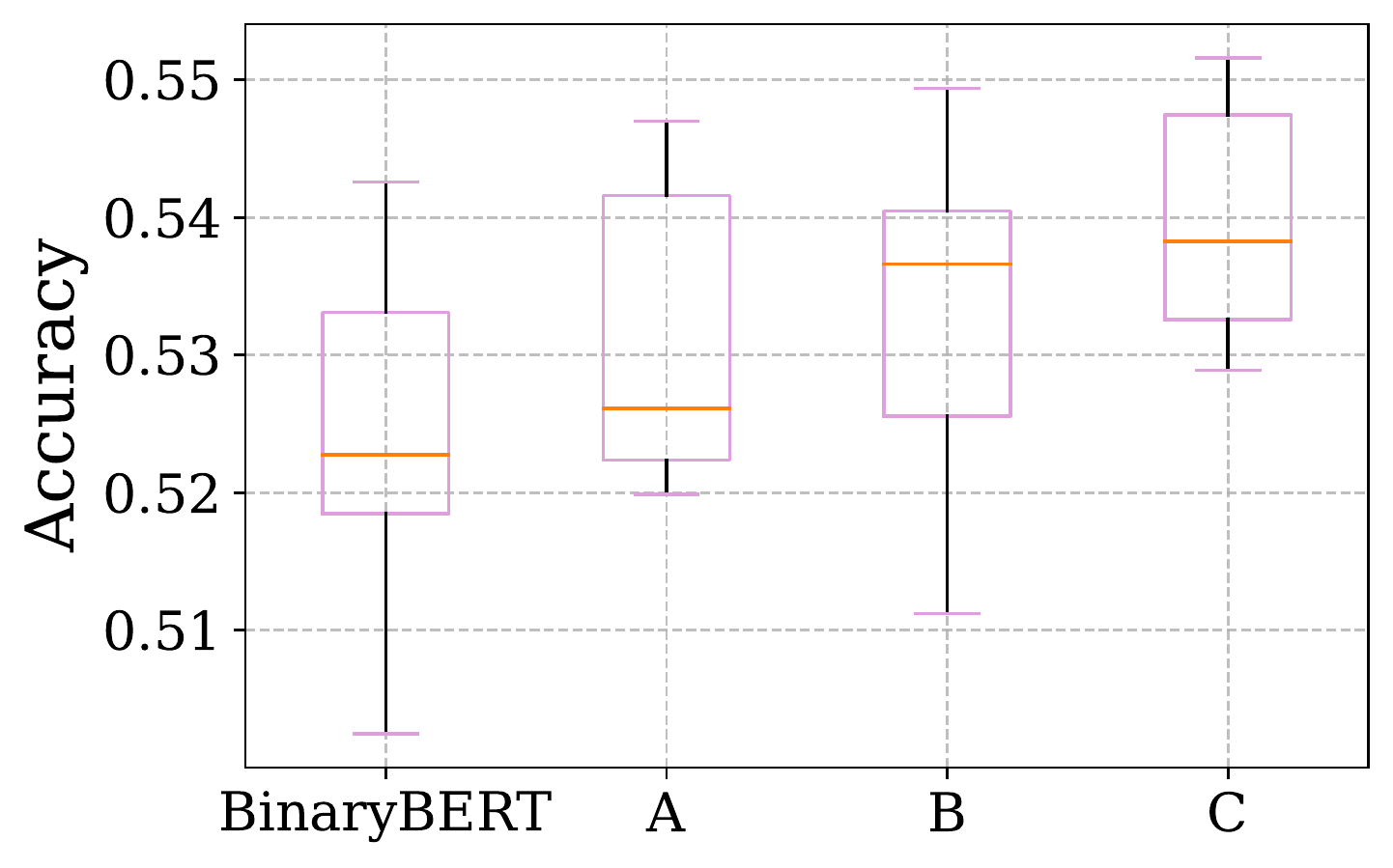}} 
\subfloat[SST-2]{  
\centering  
\includegraphics[width=0.5\linewidth,scale=1.00]{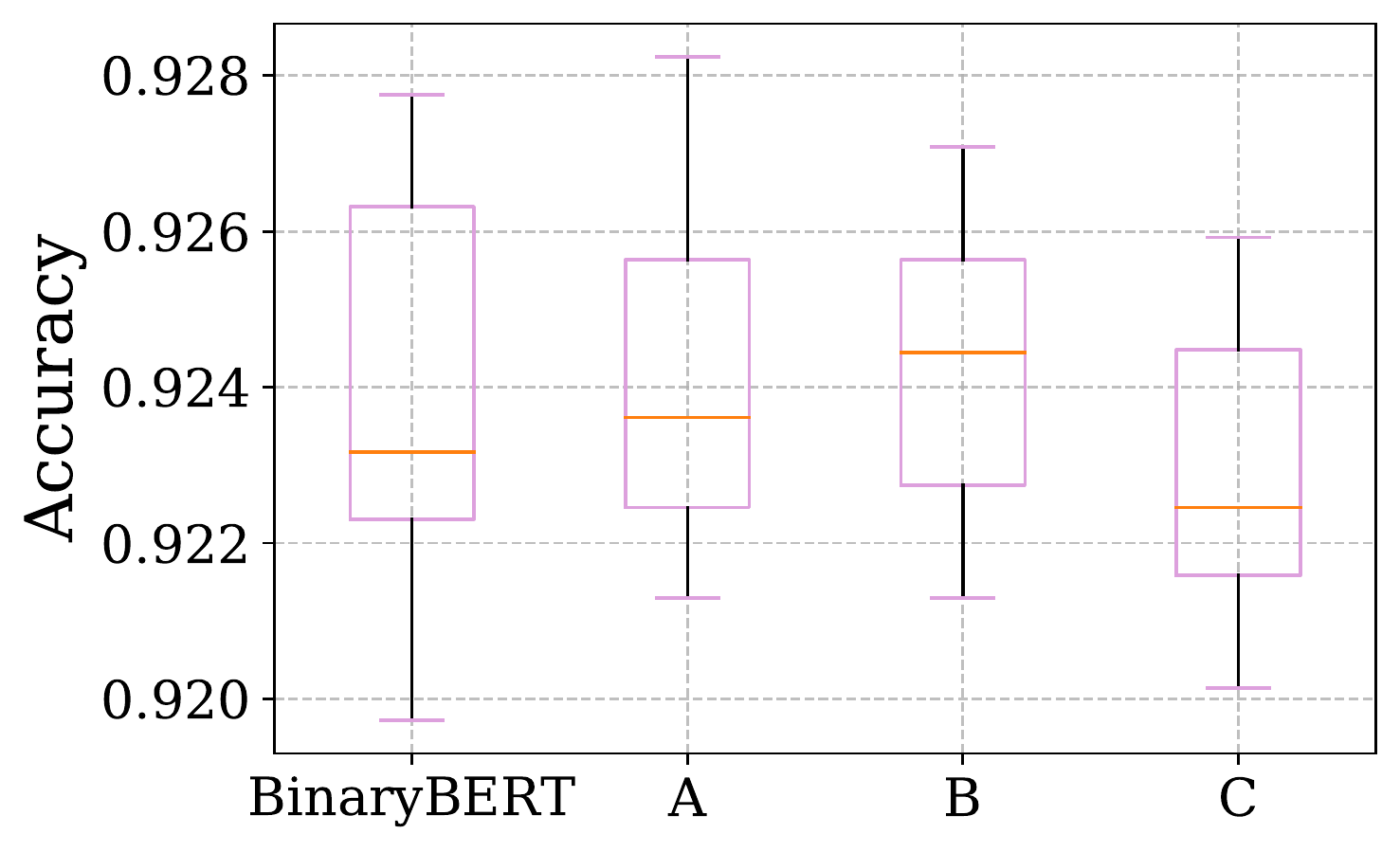}}  
\caption{Ablation study for BEBERT with different KD methods (A: w/o KD, B: prediction KD, C: two-stage KD).}
\label{fig:3} 
\vspace{-1.2em}
\end{figure}

\Rone{Inspired by the empirical opinion in \cite{allen2020towards} that convolutional neural networks can improve little accuracy if using ensemble learning after the KD procedures, we remove the KD during ensemble for accelerating the training of BEBERT.}
Although the two-stage KD has better performances in \cite{jiao2020tinybert}, it is time-consuming to conduct forward and backward propagation twice. 
Ensemble with prediction KD can avoid double propagation and ensemble can even remove the evaluation process of the teacher model. 
Fig. \ref{fig:3} shows whether applying KD in ensemble BinaryBERT has a minor effect on its accuracy in the GLUE datasets, showing that our BEBERT without KD can save training time while preserving accuracy.

\begin{table*}[htbp]
\centering
\resizebox{0.88\textwidth}{!}{
\begin{tabular}{l|l|l|l|lllllll|l}
\hline
\textbf{Method}                   & \textbf{\#Bits (W-E-A)} & \textbf{KD} & \textbf{DA} & \textbf{MNLI-m/mm} & \textbf{QQP}  & \textbf{QNLI}  & \textbf{SST-2} & \textbf{CoLA} & \textbf{MRPC} & \textbf{RTE}  & \textbf{Avg. (\%)}  \\ \hline
\textbf{BiBERT}                  & 1-1-1           & \textbf{D}  & \XSolidBrush  & 66.1/67.5             & 84.8          & 72.6           & 88.7           & 25.4          & 72.5          & 57.4          & 66.99          \\ 
\textbf{BEBERT}                  & 1-1-1           & \textbf{D}  & \XSolidBrush  & \textbf{75.7/76.6}    & \textbf{84.9} & \textbf{80.7}  & \textbf{90.2}  & \textbf{27.7} & \textbf{75.1} & \textbf{58.6} & \textbf{70.54} \\ \hline
\textbf{BiBERT}                  & 1-1-1           & \textbf{D}  & \Checkmark  & 66.1/67.5             & 84.8          & 76.0             & 90.9           & 37.8          & 78.8          & 61.0            & 70.97          \\ 
\textbf{BEBERT}                  & 1-1-1           & \textbf{D}  & \Checkmark  & \textbf{75.7/76.6}    & \textbf{84.9} & \textbf{80.7} & \textbf{91.8}  & \textbf{42.6} & \textbf{82.4} & \textbf{65.8} & \textbf{74.97} \\ \hline
\textbf{BinaryBERT}              & 1-1-4           & \textbf{C}  & \XSolidBrush  & 83.9/84.2          & 91.2          & 90.9          & 92.3           & 44.4          & 83.3          & 65.3          & 78.76          \\ \hdashline
{\textbf{BEBERT}} & 1-1-4           & \textbf{A}  & \XSolidBrush  & 84.7/84.6          & 91.3          & \textbf{91.2} & \textbf{93.3}  & 52.2          & \textbf{85.2} & \textbf{68.8} & 80.96          \\
                                 & 1-1-4           & \textbf{B}  & \XSolidBrush  & \textbf{84.7/84.8} & 91.2          & \textbf{91.2} & 93.1           & 52.2          & \textbf{85.2} & \textbf{68.8} & 80.91          \\
                                 & 1-1-4           & \textbf{C}  & \XSolidBrush  & 84.7/84.7          & \textbf{91.4} & 91.1          & 93.2           & \textbf{54.0}   & 84.6          & \textbf{68.8} & \textbf{81.11} \\ \hline
\textbf{BinaryBERT}              & 1-1-4           & \textbf{C}  & \Checkmark  & 83.9/84.2          & 91.2          & 91.4          & \textbf{93.7}           & 53.3          & \textbf{86.0}   & 71.5          & 81.57          \\ \hdashline
{\textbf{BEBERT}} & 1-1-4           & \textbf{A}  & \Checkmark  & 84.7/84.6          & 91.3          & \textbf{91.6}              & 93.4               & \textbf{58.3} & 85.7          & \textbf{72.7} & \textbf{82.53}               \\
                                 & 1-1-4           & \textbf{B}  & \Checkmark  & \textbf{84.7/84.8} & 91.2          & 91.5 & 93.1           & \textbf{58.3} & 85.7          & \textbf{72.7} & 82.46 \\
                                 & 1-1-4           & \textbf{C}  & \Checkmark  & 84.7/84.7          & \textbf{91.4} & 91.5 & 93.4  & 56.5          & 85.2          & 71.5          & 82.03          \\ \hline
\textbf{baseline}                & 32-32-32        & \textbf{-}  & \textbf{-}  & 84.9/85.5               & 91.5          & 92.1          & 93.2           & 59.7          & 86.3          & 72.2          & 82.84          \\ \hline
\end{tabular}}
\caption{Results on the GLUE development set. “\#Bits (W-E-A)” represents the bitwidth for weights of Transformer layers, word embedding, and activations. Note that there are four KD strategies for BEBERTs (A: w/o KD, B: prediction KD, C: two-stage KD, D: DMD).
}
\label{table:1}
\vspace{-1.5em}
\end{table*}

\vspace{-0.5em}
\section{Experimental Results} \label{sec:res}
\subsection{Experimental Setup} 
\label{sec: ex setup}
We evaluate BEBERT based on BinaryBERT \cite{bai2021binarybert} and BiBERT \cite{qin2022bibert} on the GLUE benchmark \cite{wang2018glue}.
\Ree{The number of ensemble models is 2 for BinaryBERT-based BEBERT and 8 for BiBERT-based BEBERT, respectively.}
\Rone{Following prior works \cite{hou2020dynabert, jiao2020tinybert, zhang2020ternarybert, bai2021binarybert}, we train BinaryBERT-based BEBERT using 3 epochs without DA on MNLI and QQP, while for other tasks we train for 1 epoch with DA and perform vanilla training for 6 epochs.}
Following the training settings of BiBERT, we train ensemble BiBERT for 50 epochs on CoLA, 20 epochs on MRPC, STS-B, and RTE, 10 epochs on SST-2 and QNLI, and 5 epochs on MNLI and QQP\Rone{, respectively}.
If KD is employed, BEBERT takes a full-precision finetuned DynaBERT \cite{hou2020dynabert} as the teacher model. 

\vspace{-0.7em}
\subsection{Analysis of Model Accuracy and Robustness} 
\label{sec: acc rob}
We evaluate the accuracy of BEBERTs on the GLUE development set, and the experimental results are shown in Table~\ref{table:1}.
\Rone{Our BiBERT-based and BinaryBERT-based BEBERTs improve accuracy by 4.0\% and 2.4\% on average, respectively, in comparison with BiBERT and BinaryBERT.}
\Rone{Especially for the CoLA task, BiBERT-based and BinaryBERT-based BEBERTs with DA increase accuracy by 4.8\% and 5\%, respectively.}
Moreover, BiBERT (W-E-A: 1-1-1) attains more accuracy increase on average after ensemble compared with BinaryBERT (W-E-A: 1-1-4), showing the great potential of our approach applying for extremely low-bit quantized models.

\begin{figure}[htbp]  
\centering  
\includegraphics[width=1.0\linewidth]{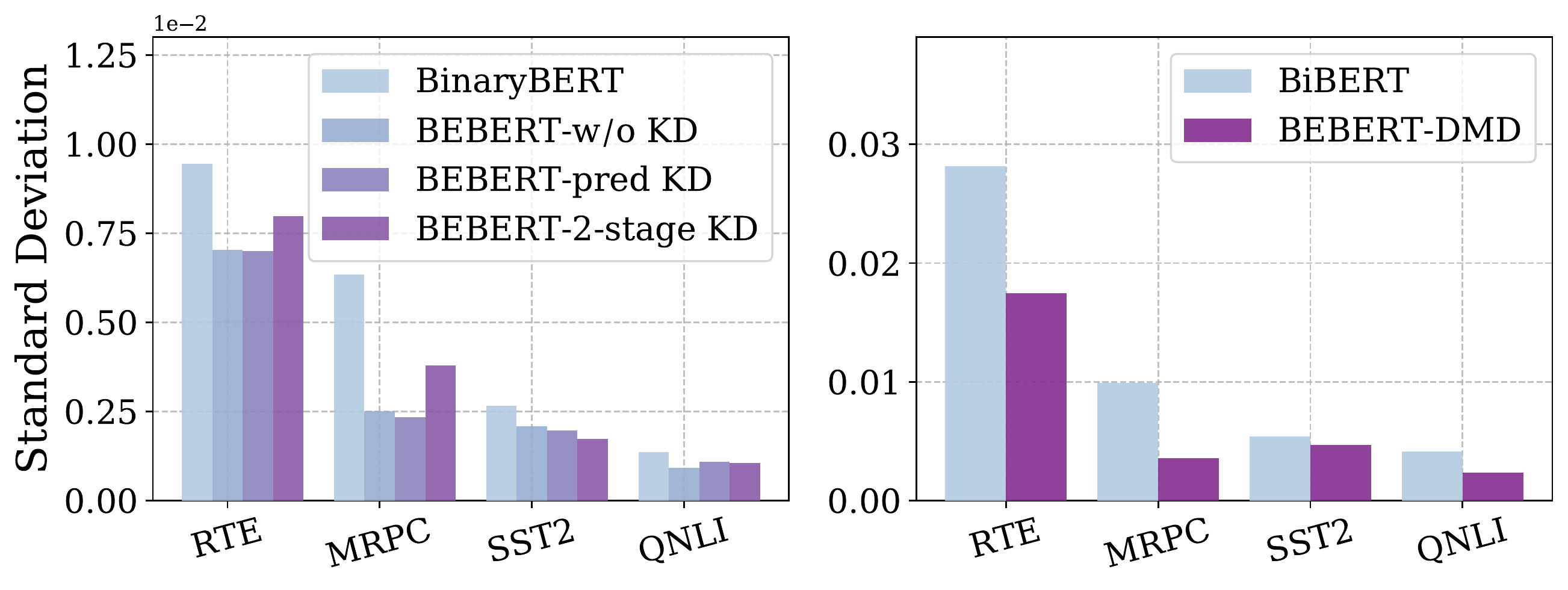}
\caption{Robustness analysis for BEBERT w/o DA. \textbf{Left}: ensemble BinaryBERT. \textbf{Right}: ensemble BiBERT.} 
\label{fig:4} 
\vspace{-1.2em}
\end{figure}

To \Rone{evaluate the robustness} of BEBERTs, we inject the input perturbation $\Delta x$ in the embedding layer by Gaussian noise with a variance of 0.01 \Rone{following the steps in} \cite{hua2021noise}.
As shown in Fig.~\ref{fig:4}, the standard deviation of the output accuracy in BinaryBERT-based and BiBERT-based BEBERT are reduced by 65\% and 61\%, respectively, compared with their baselines, indicating considerable stability improvement.

Although ensemble techniques improve the accuracy and robustness of BiBERT, the experimental results indicate that the BiBERT-based BEBERT is impractical for deployment. Therefore, we focus on the evaluation of BinaryBERT-based BEBERT in the following experiments.

\vspace{-0.7em}
\subsection{Analysis of Training Efficiency} 
\begin{figure}[htbp]  
\centering  
\includegraphics[width=0.96\linewidth]{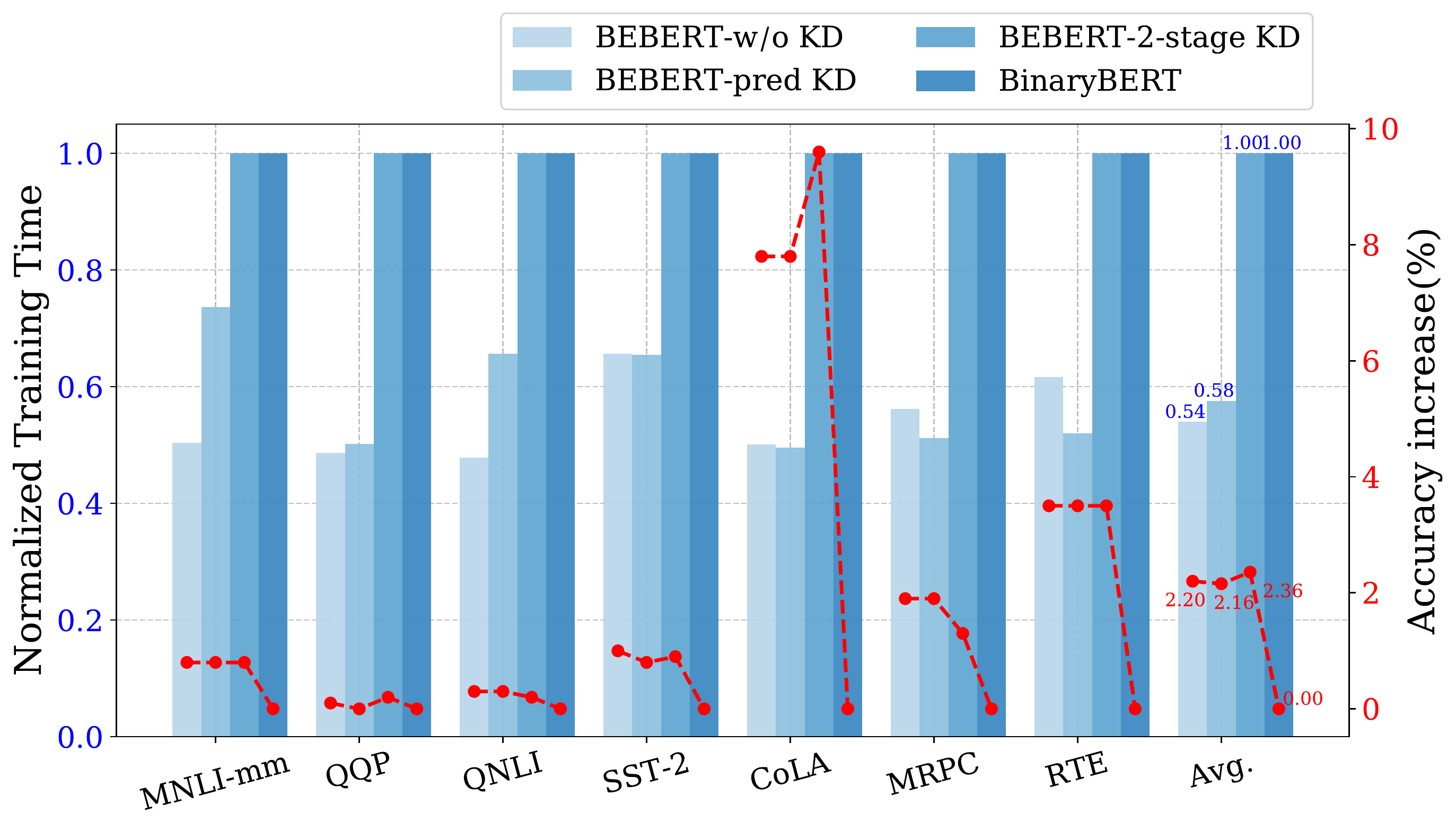}
\caption{The required training time of BEBERT w/o DA using different KD methods.}
\label{fig:5} 
\vspace{-1.3em}
\end{figure}

We evaluate the training efficiency of BEBERT w/o KD compared to BEBERTs trained with various KD methods. 
\Ree{Theorically, the two-stage KD used in our BinaryBERT baseline brings two forward and backward propagation processes, which could double the training time. Additionally, if we remove the KD process, we could remove the full-precision teacher during training, which brings an impressive reduction in memory use.}
Fig.~\ref{fig:5} shows the normalized training time and accuracy increase of BEBERT.
Our BEBERT w/o KD achieves a 2.2\% accuracy increase with \Ree{approximately} 2× speedup during the training process, which demonstrates the efficiency of removing the KD procedure during training.
\vspace{-0.5em}

\vspace{-0.7em}
\subsection{Comparison with the SOTAs}
\Rone{To understand the effectiveness of BEBERT for deployment, we compare BEBERT to various SOTA compressed BERTs using quantization \cite{shen2020q, bai2021binarybert, zhang2020ternarybert}, pruning \cite{chen2021earlybert, hou2020dynabert}, KD \cite{wang2020minilm, sanh2019distilbert}, and DA \cite{jiao2020tinybert} in terms of model accuracy and compression ratio.}
Inference of BEBERT can be performed in parallel due to the parallelism potential of binary models \cite{li2019bstc, zhao2017accelerating}, avoiding the increase in FLOPs. 
BEBERT outperforms the SOTA compressed BERTs in accuracy by up to 6.7\%.
Compared to the full-precision BERT, it also saves 15× and 13× on FLOPs and model size, respectively, with a negligible accuracy loss of 0.3\%, showing the potential for practical deployment. 

\begin{table}[]
\centering
\resizebox{0.44\textwidth}{!}{
\begin{tabular}{l|ccc|c||c}
\hline
\multicolumn{1}{c|}{\textbf{Method}} & \textbf{\begin{tabular}[c]{@{}c@{}}\#Bits\\ (W-E-A)\end{tabular}} & \textbf{\begin{tabular}[c]{@{}c@{}}Size\\ (MB)\end{tabular}} & \textbf{\begin{tabular}[c]{@{}c@{}}FLOPs\\ (G)\end{tabular}} & \textbf{DA} & \textbf{\begin{tabular}[c]{@{}c@{}}Avg.\\(\%)\end{tabular}}  \\ \hline
\textbf{EarlyBERT \cite{chen2021earlybert}}                  & full-prec.                                                        & 133                                                          & 10.1                                                         & \XSolidBrush  & 75.80           \\
\textbf{DynaBERT \cite{hou2020dynabert}}                   & full-prec.                                                        & 33                                                           & 0.8                                  & \XSolidBrush  & 77.36          \\
\textbf{DistilBERT6L \cite{sanh2019distilbert}}               & full-prec.                                                        & 264                                                          & 11.3                                                         & \XSolidBrush  & 78.56          \\
\textbf{BinaryBERT \cite{bai2021binarybert}}                 & 1-1-4                                                             & 16.5                                                         & 1.5                                                          & \XSolidBrush  & 78.76          \\
\textbf{Q-BERT \cite{shen2020q}}                     & 8-8-8                                                             & 104                                                          & 5.6                                                          & \XSolidBrush  & 79.83          \\
\textbf{miniLM \cite{wang2020minilm}}                     & full-prec.                                                        & 264                                                          & 11.3                                                         & \XSolidBrush  & 80.71          \\
\textbf{BEBERT}                     & 1-1-4                                                             & 33                                                           & 3.0/2                                                          & \XSolidBrush  & \textbf{80.96} \\ \hline
\textbf{BinaryBERT \cite{bai2021binarybert}}                 & 1-1-4                                                             & 16.5                                                         & 1.5                                                          & \Checkmark  & 81.57          \\
\textbf{TinyBERT6L \cite{jiao2020tinybert}}                 & full-prec.                                                        & 264                                                          & 1.2                                                          & \Checkmark  & 81.91          \\
\textbf{TernaryBERT \cite{zhang2020ternarybert}}                & 2-2-8                                                             & 28                                                           & 6.0                                                            & \Checkmark  & 81.91          \\
\textbf{BEBERT}                     & 1-1-4                                                             & 33                                                           & 3.0/2                                                          & \Checkmark  & \textbf{82.53} \\ \hline
\textbf{baseline}                   & full-prec.                                                        & 418                                                          & 22.5                                                         & \textbf{-}  & 82.84          \\ \hline
\end{tabular}}
\caption{Comparison with the SOTA compressed BERTs on the GLUE development set.}
\vspace{-1.2em}
\end{table}

\section{Conclusion}

In this paper, an efficient and robust binary ensemble BERT model, namely BEBERT, is proposed for practical deployment \R{targeting at} the resource-constrained edge devices.
\R{For the first time, we} introduce ensemble techniques to binary BERT models, boosting the accuracy and robustness while retaining the computational efficiency. 
Moreover, we accelerate the training process of BEBERT by removing redundant knowledge distillation processes during ensemble without impact on accuracy.
Experimental results on the GLUE benchmark show BEBERT \R{with efficient training} significantly outperforms the binary BERTs in accuracy and robustness, and \R{exceeds} the state-of-the-art compressed BERTs \R{in accuracy}.


\clearpage
\bibliographystyle{IEEEbib}
\bibliography{refs}

\end{document}